\documentclass[letterpaper, 10 pt, journal, twoside]{ieeetran}  

\IEEEoverridecommandlockouts                              

\makeatletter
\let\NAT@parse\undefined
\makeatother

\usepackage[dvipsnames]{xcolor}
\usepackage{multicol}
\usepackage{graphicx}
\usepackage{float}
\usepackage{booktabs}
\usepackage{amsmath}
\usepackage{amssymb}
\usepackage{algorithm}
\usepackage{algpseudocode}
\usepackage{cuted}
\usepackage{capt-of}  
\usepackage[caption=false]{subfig}
\usepackage{seqsplit}

\usepackage[bookmarks=true]{hyperref}
\hypersetup{
	colorlinks=true,
	linkcolor=red,
	filecolor=magenta,      
	urlcolor=red,
	citecolor=red,
}

\begin{document}

\title{Squint: Fast Visual Reinforcement Learning for Sim-to-Real Robotics}

\newcommand{\website}{\url{https://aalmuzairee.github.io/squint}}

\markboth{IEEE Robotics and Automation Letters. Preprint Version. Accepted August, 2026}
{Almuzairee \MakeLowercase{\textit{et al.}}: Squint: Fast Visual Reinforcement Learning for Sim-to-Real Robotics} 

\author{Abdulaziz Almuzairee$^{1}$ and Henrik I. Christensen$^{1}$%
  \\
  \vspace{0.1cm}
  \large\website

\thanks{Manuscript received: April, 28, 2026; Revised July, 29, 2026; Accepted August, 20, 2026.}
\thanks{This paper was recommended for publication by Editor Jens Kober upon evaluation of the Associate Editor and Reviewers’ comments.
This work in part was supported by Kuwait University through a visiting fellow program.} 
\thanks{$^{1}$The authors are with Department of Computer Science and Engineering, University of California San Diego, La Jolla, CA 92093, USA 
        {\tt\footnotesize \{aalmuzairee,hichristensen\}@ucsd.edu}}%
\thanks{For code and videos: \url{https://aalmuzairee.github.io/squint}}
\thanks{Digital Object Identifier (DOI): see top of this page.}

}

\urlstyle{tt}  

\maketitle

\begin{strip}
    \centering
    
    \vspace{-3.2cm}
    \includegraphics[width=\textwidth]{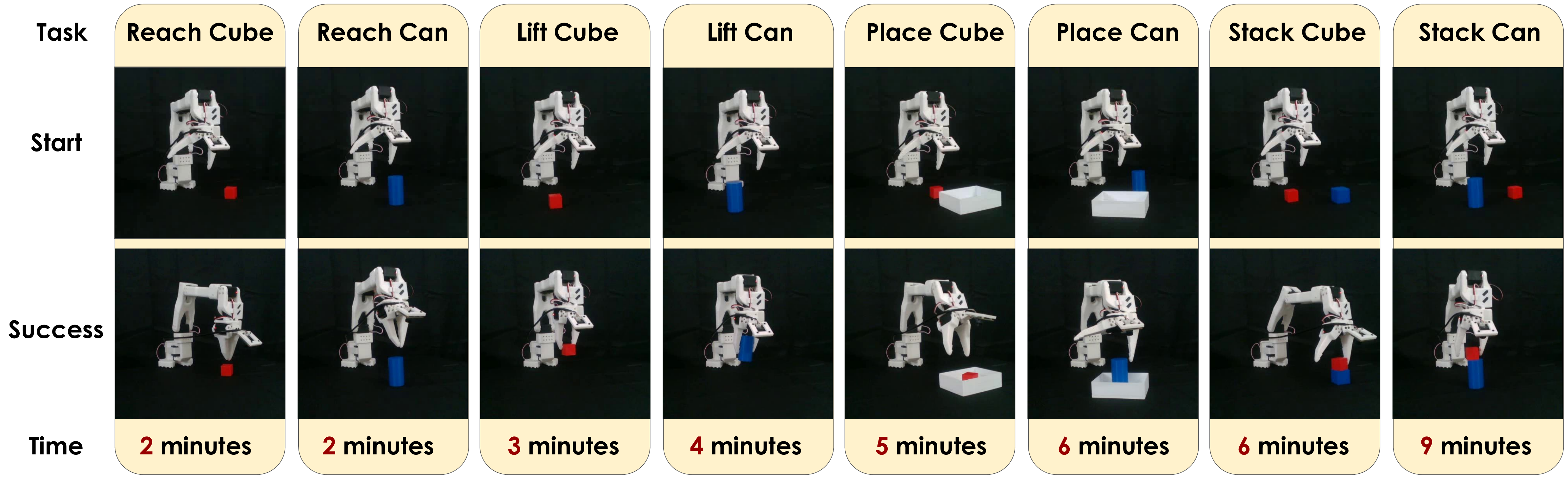}
    \vspace{-0.1cm}
    
    \refstepcounter{figure}
    {\small Fig.~\thefigure. \textbf{Fast Sim-to-Real Robotics.} We setup 8 visual tasks on the 5 DoF SO-101 Robotics Arm in ManiSkill3 and train our method, \emph{Squint}, purely from images and proprioceptive state on these tasks for 15 minutes. After 15 minutes, we deploy the final checkpoint zero-shot to our real robot setup. We show visuals of the real robot for each of the tasks above, along with an image of the "Start" frame, the final "Success" frame, and the "Time" it takes our agent to converge in simulation for each task on a single 3090 RTX GPU.}
    \label{fig:header}
\end{strip}

\begin{abstract}

Visual reinforcement learning is appealing for robotics but expensive. Off-policy methods are sample-efficient yet slow while on-policy methods parallelize well but waste samples. Recent work has shown that off-policy methods can train faster than on-policy methods in wall-clock time for state-based control. Extending this to vision remains challenging, where high-dimensional input images complicate training dynamics and introduce substantial storage and encoding overhead. To address these challenges, we introduce Squint, a visual Soft Actor Critic method that achieves faster wall-clock training than prior visual off-policy and on-policy methods. Squint achieves this via parallel simulation, a distributional critic, resolution squinting, layer normalization, a tuned update-to-data ratio, and an optimized implementation. We evaluate on the SO-101 Task Set, a new suite of eight manipulation tasks in ManiSkill3 with heavy domain randomization, and demonstrate sim-to-real transfer to a real SO-101 robot. We train policies for 15 minutes on a single RTX 3090 GPU, with most tasks converging in under 6 minutes.

\end{abstract}



\section{Introduction}
\label{sec:into}

\begin{figure*}[t]
    \centering
    \includegraphics[width=1.0\linewidth,trim={0.cm 0cm 0.0cm 0cm},clip]{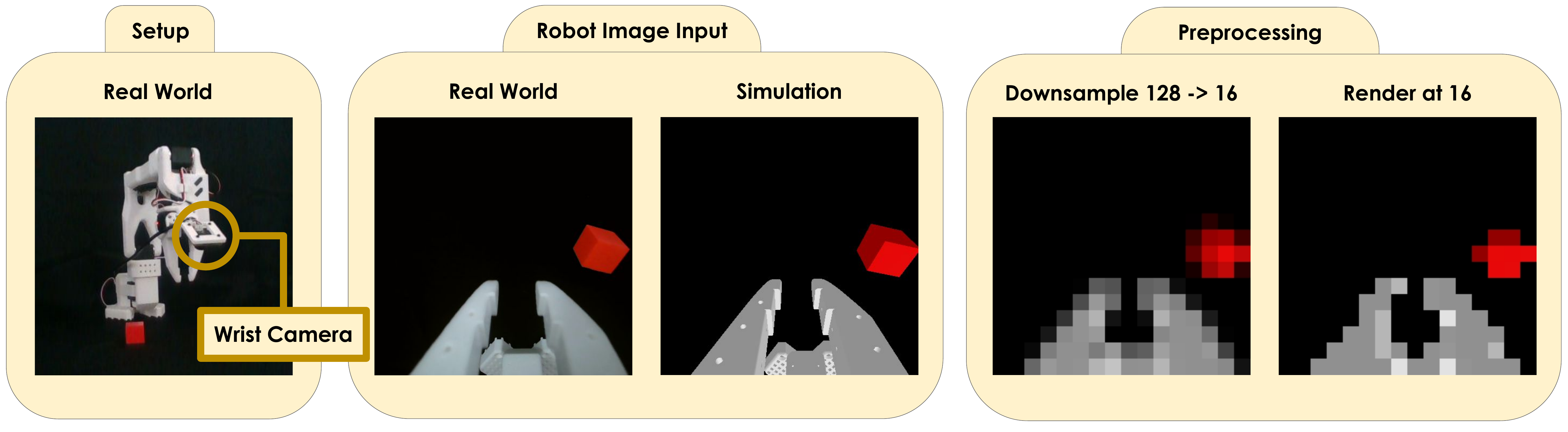}
    
    \vspace{-0.1cm}

    \refstepcounter{figure}
    \begin{flushleft}
    {\small Fig.~\thefigure. \textbf{Robot Setup.} \textit{(Left)} Setup of the real robot from an external camera. \textit{(Middle)} The input camera image for the robot in both the real world and simulation. We only use the wrist camera as the input image for the robot. \textit{(Right)} We display different preprocessing effects, downsampling from a higher resolution vs rendering at the target resolution directly.}
    \end{flushleft}

    \vspace{-0.5cm}

   \label{fig:setup}
\end{figure*}

\IEEEPARstart{V}{isual} reinforcement learning (RL) \cite{mnih2013playing} offers a compelling paradigm for robotics, enabling deployment onto real robots with camera inputs and requiring no task-specific instrumentation \cite{kalashnikov2018qt, zhuang2023robot, kaufmann2023drone}. Despite this appeal, training visuomotor policies remains notoriously expensive, demanding millions of environment interactions and substantial compute.

Over the past few years, researchers have made strides in reducing the number of environment interactions needed, often referred to as sample efficiency, by improving off-policy algorithms. These off-policy algorithms, such as SAC \cite{haarnoja2018soft}, TD3 \cite{fujimoto2018addressing}, and TD-MPC2 \cite{hansen2024tdmpc2}, reuse experiences through a replay buffer. Yet, these improvements come with computational costs that increase training wall-time of these agents in simulation. This slows research iteration. On the other hand, on-policy methods like Proximal Policy Optimization \cite{schulman2017proximal} offer poor sample efficiency, without any experience reuse, but can parallelize trivially across thousands of environments on modern GPUs, achieving faster wall-times than their off-policy counterparts. This has led to on-policy dominance in GPU-accelerated simulators and PPO to be the de facto standard for sim-to-real robotics in the past decade \cite{heess2017emergence, hwangbo2019learning}.

While sample efficiency and wall-time are both important axes for reinforcement learning, recent works like FastTD3 and FastSAC \cite{seo2025fasttd3, seo2025learning} explore an intermediate axis, one where off-policy algorithms can be tuned to maximize wall-time rather than sample efficiency, achieving faster wall times than their on-policy counterparts in parallelized simulations. These works made significant progress in scaling off-policy learning for state-based control. Yet, extending these works to visual reinforcement learning remains nontrivial: high-dimensional input images complicate training dynamics, storing them in replay buffers strains memory, and encoding them through convolutional networks adds computational overhead. 

Motivated by these challenges, we introduce our method, \emph{Squint}, a visual based Soft Actor Critic method, that through careful image preprocessing, architectural design choices, and hyperparameter selection, is able to leverage parallel environments and experience reuse effectively, achieving faster wall-clock training time than both prior visual off-policy and on-policy methods, and solving visual robotic tasks in minutes. 

To train Squint, we leverage ManiSkill3 \cite{tao2024maniskill3}, a fast-GPU simulator, with powerful batched rendering capabilities, and create a digital twin of our real robot setup. Our real robot setup consists of a 5 DoF SO-101 Robot Arm \cite{cadene2024lerobot}. We build 8 distinct tasks, which we term the \emph{SO-101 Task set}, and use them as a testbed for investigating scalable visual reinforcement learning. We train our method and baselines on the SO-101 Task set for 15 minutes, and deploy them to our real robot setup, as shown in Figure \ref{fig:header} and Figure \ref{fig:setup}.

Our key contributions are \textbf{(1)} Squint: A fast off-policy visual actor critic method that achieves faster wall-clock training than prior visual RL methods, with extensive experiments justifying each design choice. \textbf{(2)} SO-101 Task Set: 8 manipulation tasks in ManiSkill3 with heavy domain randomization for sim-to-real transfer. \textbf{(3)} Real-world validation: Policies trained in 15 minutes on a single RTX 3090, deployed zero-shot to the real-world SO-101. 
\emph{To the best of our knowledge, Squint is the first method to demonstrate zero-shot sim-to-real transfer of a contact-rich stacking policy, trained entirely from scratch with no demonstrations, in under 15 minutes on a single consumer GPU. }

\section{Related Work}
\label{sec:related}

\textbf{Parallel Simulators for Robotics.} 
Simulators provide a fast way for reinforcement learning agents to learn in a safe, constrained environments \cite{bellemare2013arcade, collins2021review}. Early parallelization in CPU allowed agent to train faster in terms of wall time \cite{heess2017emergence, tassa2018deepmind, kalashnikov2018qt, yu2020meta, zhu2020robosuite}.
With the advent of fast GPU-based parallelization, these simulators became faster, and allowed the community to iterate faster on training and deployment. \cite{liang2018gpu,makoviychuk2021isaac, mu2021maniskill, gu2023maniskill2, tao2024maniskill3, sferrazza2024humanoidbench, zakka2025mujoco} Most notably, ManiSkill3 \cite{tao2024maniskill3} provides fast GPU-based batched rendering, which allows researchers to train fast visual based reinforcement learning agents for deployment. Due to its realism, and ease of use, we use ManiSkill3 as our main simulator for this work.  Despite its realism, domain randomization \cite{tobin2017domain} is necessary to bridge the sim-to-real gap. Thus, we apply domain randomization to our task set to allow better sim to real transfer.

\textbf{Visual Reinforcement Learning.} Learning from vision has driven the start of deep RL in the recent decade \cite{mnih2013playing, mnih2015human}. Building on it, many works explored different methods to improve off-policy training sample efficiency, including: autoencoder architectures \cite{finn2015learning, yarats2019improving}, contrastive learning representations \cite{srinivas2020curl}, data regularized representations \cite{laskin2020reinforcement, yarats2021mastering}, and model-based representations \cite{hafner2023dreamer, hansen2024tdmpc2, fujimoto2025mrq}. Perhaps most notably, DrQ-v2 \cite{yarats2021mastering} is widely used as a solid baseline for visual reinforcement learning sample efficiency. Unfortunately, this sample efficiency comes at the cost of training wall-time, which on-policy PPO \cite{schulman2017proximal} exceeds at in scale. While most prior visual off policy learning algorithms focus on sample efficiency, we follow PQL\cite{li2023parallel}, PQN \cite{gallici2024simplifying}, FastTD3 \cite{seo2025fasttd3} and FastSAC \cite{seo2025learning} in focusing on training wall-time, and introduce our algorithm, Squint, to excel at this axis in the visual RL domain.

\begin{figure*}[t]
    \centering
    \includegraphics[width=0.9\linewidth,trim={0.cm 0.2cm 0.0cm 0.1cm},clip]{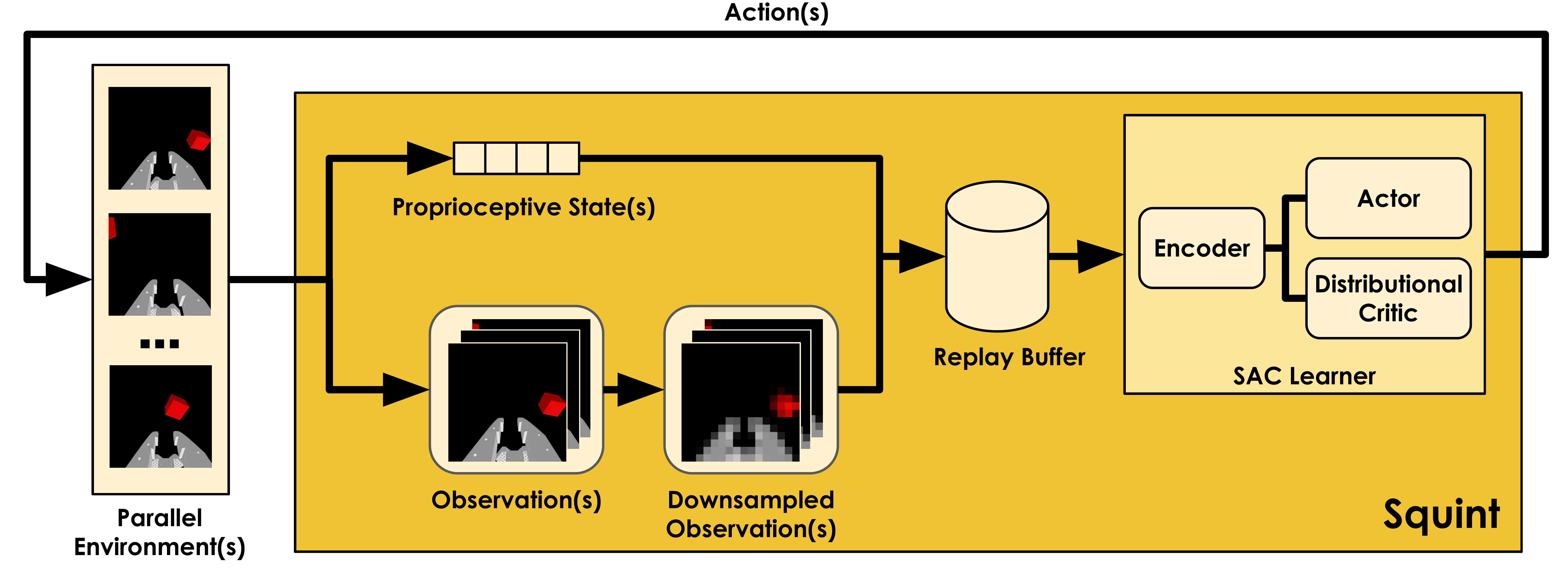}
    
    \vspace{-0.1cm}

    \refstepcounter{figure}
    \begin{flushleft}
    {\small Fig.~\thefigure. \textbf{Method Diagram.} Squint is a fast off-policy actor-critic visual reinforcement learning algorithm that accelerates training speed by leveraging parallel environments and downsampling input image observations. }
    \end{flushleft}

    \vspace{-0.5cm}

   \label{fig:method_diagram}
\end{figure*}

\section{Background}
\label{sec:background}

\textbf{Visual Reinforcement Learning} formulates the interaction between the agent and the environment as a Partially Observable Markov Decision Process (POMDP) \cite{kaelbling1998pomdp}, defined by a tuple of $\langle \mathcal{S}, \mathcal{O}, \mathcal{A}, \mathcal{T}, R, \gamma \rangle$. In a POMDP, ${s} \in \mathcal{S}$ is the state space, ${o} \in \mathcal{O}$ is the observation space, ${a} \in \mathcal{A}$ the action space, $\mathcal{T}: \mathcal{S} \times \mathcal{A} \times \mathcal{S} \rightarrow [0, 1]$ is the state-to-state transition probability function, $R: \mathcal{S} \times \mathcal{A} \rightarrow \mathbb{R}$ is the reward function and $\gamma$ is the discount factor. At each time step $t$, the agent receives as input: a single RGB image as an observation ${o} \in \mathcal{O}$ and \emph{optionally} the proprioceptive state $s^{proprio} \in \mathcal{S}$, and is queried to select an action ${a}$ . The goal is to learn a policy $\pi: \mathcal{O} \rightarrow \mathcal{A}$ that maximizes the expected discounted sum of rewards $\mathbf{E}_\pi [\sum_{t=0}^\infty \gamma^t r_t]$ where $r_t =  R({s}_t, {a}_t)$.

\textbf{Soft Actor Critic} (SAC) \cite{haarnoja2018soft} is an off-policy actor critic algorithm, commonly used for continuous control. SAC concurrently learns a Q-function $Q_\theta$ (critic) and a policy $\pi_\phi$ (actor) with a separate neural network for each. The Q-function $Q_{\theta}$ aims to estimate the optimal soft state-action value function by minimizing the one step Bellman Residual:
\begin{align}
    y &= r_t + \gamma \Big( Q_{\overline{\theta}}(\mathbf{s}_{t+1}, \tilde{\mathbf{a}}_{t+1}) \nonumber \\
      &\quad - \alpha \log \pi_\phi(\tilde{\mathbf{a}}_{t+1} | \mathbf{s}_{t+1}) \Big) \\
    \mathcal{L}_{Q_{\theta}}(\mathcal{D}) &= \mathbb{E}_{(\mathbf{s}_{t},\mathbf{a}_{t}, r_{t}, \mathbf{s}_{t+1})\sim\mathcal{D}} \left[ \left( Q_\theta(\mathbf{s}_{t}, \mathbf{a}_{t}) - y \right)^2 \right]
\end{align}
where $\mathcal{D}$ is the replay buffer, $\tilde{\mathbf{a}}_{t+1} \sim \pi_\phi(\cdot | \mathbf{s}_{t+1})$, $Q_{\overline{\theta}}$ represents an exponential moving average of the weights from $Q_\theta$, and $\alpha$ is the temperature parameter \cite{watkins1992q,lillicrap2015continuous,haarnoja2018soft}. SAC also employs clipped double-Q learning \cite{Hasselt2016CDQ, fujimoto2018addressing} which we omit from our notation for simplicity. The policy $\pi_\phi$ is a stochastic policy that aims to maximize both entropy and $Q$-values:
\begin{equation}
    \mathcal{L}_{\pi_\phi}(\mathcal{D}) = \mathbb{E}_{\mathbf{s}_{t} \sim \mathcal{D}} \left[ \alpha \log \pi_\phi(\tilde{\mathbf{a}}_t | \mathbf{s}_t) - Q_\theta(\mathbf{s}_t, \tilde{\mathbf{a}}_t) \right]
\end{equation}
where $\tilde{\mathbf{a}}_t \sim \pi_\phi(\cdot | \mathbf{s}_t)$. The temperature parameter $\alpha$ can either be fixed or learned by minimizing against a target entropy $\mathcal{H}^{\text{target}}$ \cite{haarnoja2018soft}. Both the critic and actor losses are updated iteratively using stochastic gradient descent in an aim to maximize the expected discounted sum of rewards.

When learning from images, the SAC agent does not have access to state $\mathbf{s}_t$ but rather observation $\mathbf{o}_t$. We follow DrQ and DrQ-v2 \cite{kostrikov2020image, yarats2021mastering} in encoding $\mathbf{o}_t$ with a single shared CNN encoder $f_{\psi}$, such that all previous state inputs $\mathbf{s}_t$ can be replaced with $[f_{\psi}(\mathbf{o}_t)]$ or $[f_{\psi}(\mathbf{o}_t), s_t^{proprio}]$ when proprioceptive state is available. The encoder $f_{\psi}$ is trained end-to-end by the critic gradients.

\section{Method}
\label{sec:method}

We present \textit{Squint}, a visual Soft Actor Critic \cite{haarnoja2018soft} method that trains successfully within minutes through: parallel simulation, a distributional critic \cite{bellemare2017distributional}, resolution squinting, layer normalization \cite{ba2016layer}, tuned update-to-data ratio, and an optimized implementation. These trained policies transfer zero-shot to a real-world SO-101 robotic arm. We present the flow of our training pipeline in Figure \ref{fig:method_diagram}. We further show the impact of each design choice in Figure \ref{fig:design_choices} and discuss them below:

\begin{figure*}[t]
    \centering
    \includegraphics[width=1.0\linewidth,trim={0.cm 0.2cm 0.0cm 0.1cm},clip]{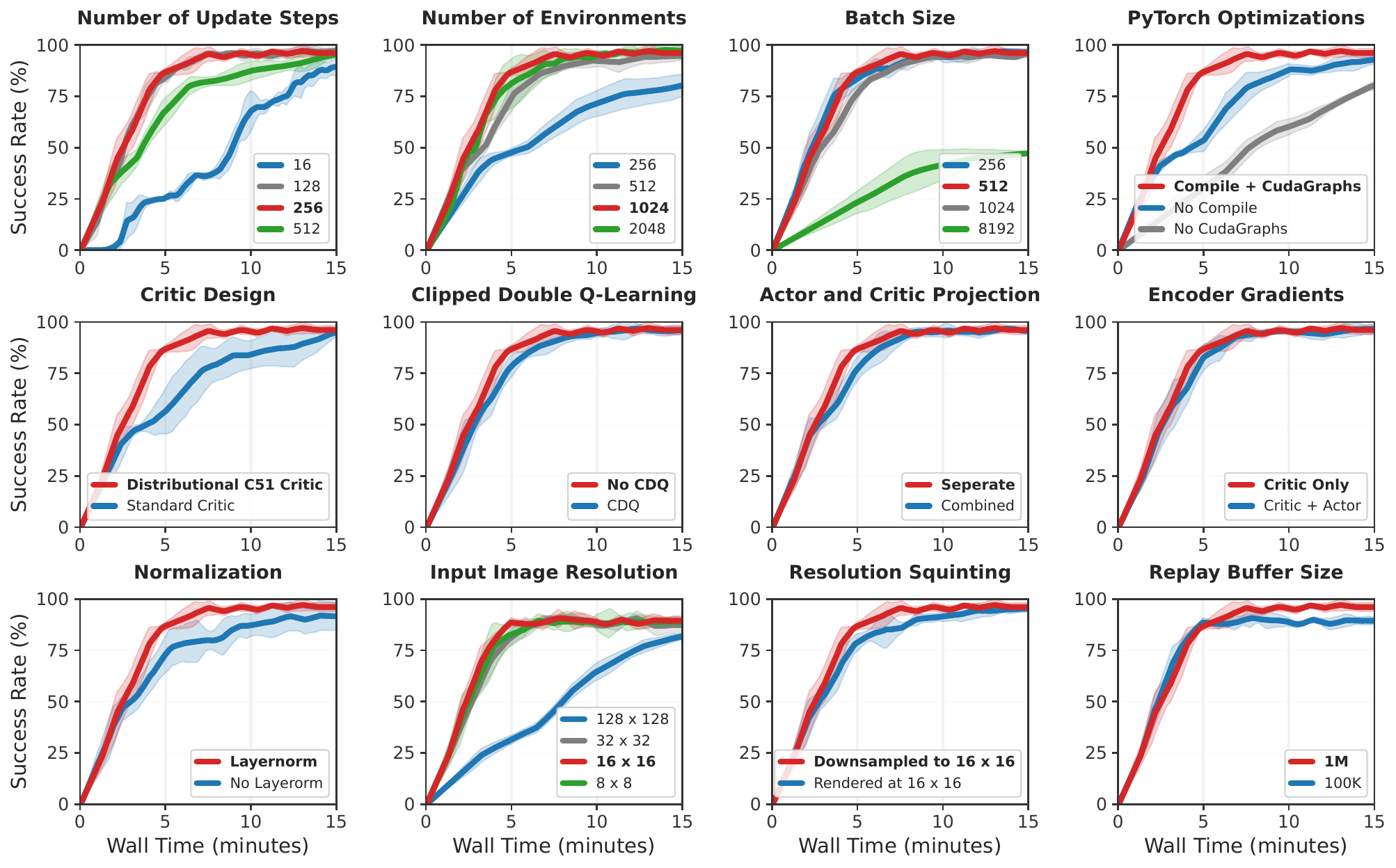}
    
    \vspace{-0.1cm}

    \refstepcounter{figure}
    \begin{flushleft}
    {\small Fig.~\thefigure. \textbf{Design Choices.} We experiment with the most critical design choices that allow our agents to be trained while maximizing wall-time for sim-to-real robotics deployment. Mean Success Rate and 95\% CI over 5 seeds for all 8 tasks.}
    \end{flushleft}

    \vspace{-0.5cm}
    
   \label{fig:design_choices}
\end{figure*}

\textbf{Update-to-Data Ratio.} Tuning the update-to-data (UTD) ratio has long been a tradeoff between sample efficiency and training wall time \cite{doro2023srsac}. For minimizing training wall time, prior work has shown that one needs to use a very large number of parallel environments with a very low number of updates \cite{shukla2025fastsac, seo2025fasttd3, seo2025learning}, creating a low UTD ratio. In Figure \ref{fig:design_choices}, we study the impact of these choices, and find that a large number of parallel environments (1024) along with a large number of updates (256) can reduce training time significantly. These findings indicate that while very low UTD ratios ($<$0.06) work for humanoids \cite{seo2025fasttd3, seo2025learning}, manipulation domains seem to benefit from larger UTD ratios, seemingly (0.25) in our task set.

\textbf{Batch Size.} While larger batch sizes have been shown to improve training efficiency in \cite{seo2025fasttd3, seo2025learning}, we find that using a lower batch size of 512 provides the same stable training efficiency, while minimizing wall time for every update. 

\textbf{Pytorch Optimizations.} Our codebase builds on LeanRL and CleanRL \cite{huang2022cleanrl}. We add support for both PyTorch compile and cudagraphs \cite{paszke2019pytorch}, which leverage kernel fusion and reduced CPU launch overhead to significantly accelerate the training loop. We further add AMP bfloat16 support for the update loops, reducing training time with convolutional neural networks. Combined, these optimizations give more than a 5x speedup in training speed.

\textbf{Critic Design Choices.} We adopt a Distributional C51 Critic \cite{bellemare2017distributional}, similar to \cite{li2023parallel, seo2025fasttd3, seo2025learning}, where we minimize cross entropy loss rather than the mean square error on TD-targets. We find that the choice of a distributional critic greatly improves wall time speed, even though it requires more computation than a standard Critic. We further investigate the effect of Clipped Double Q-learning (CDQ) \cite{fujimoto2018addressing} and find a slight improvement in using the average rather than CDQ.

\textbf{Encoder Design Choices.} We leverage a small two layer CNN encoder, and share it between the actor and critic. The encoder is only updated by the critic's TD-loss. After encoding the images, we pass them through separate one layer projections for each of the actor and critic. We adopt these design choices following prior state-of-the-art visual RL methods \cite{kostrikov2020image, yarats2021mastering}, though our ablations reveal they yield negligible performance gains on our task set.

\textbf{Normalization.} All our linear layers are followed by Layer Normalization layers \cite{ba2016layer}, which we find to improve training speed, similar to the findings of \cite{ball2023efficient, nauman2024bigger, seo2025learning}.

\textbf{Input Image Resolution.} We compare training agents with larger and lower input image resolutions, and find them to achieve identical sample efficiency, but varying wall time training speed. We only show the wall time axis in Figure \ref{fig:design_choices}. Our findings show that the smaller the resolution, the faster the wall time, and thus we use a 16 x 16 input image resolution.

\textbf{Resolution Squinting.} We compare rendering at 16 x 16 vs rendering at a high resolution of 128 x 128 and then area downsampling to 16 x 16, which we refer to as \emph{squinting}. We find that downsampling improves performance on our task set as displayed in Figure \ref{fig:setup}. This downsampling provides natural anti-aliasing and preserves scene structure. Empirically, we find that squinting cuts down training time, while achieving similar real world success rates as higher image resolutions.

\begin{figure*}[t]
  \centering
    \resizebox{0.8\linewidth}{!}{
      \begin{minipage}{0.49\linewidth}
        \centering
        \includegraphics[width=\linewidth,trim={0cm 0.2cm 0cm 0.1cm},clip]{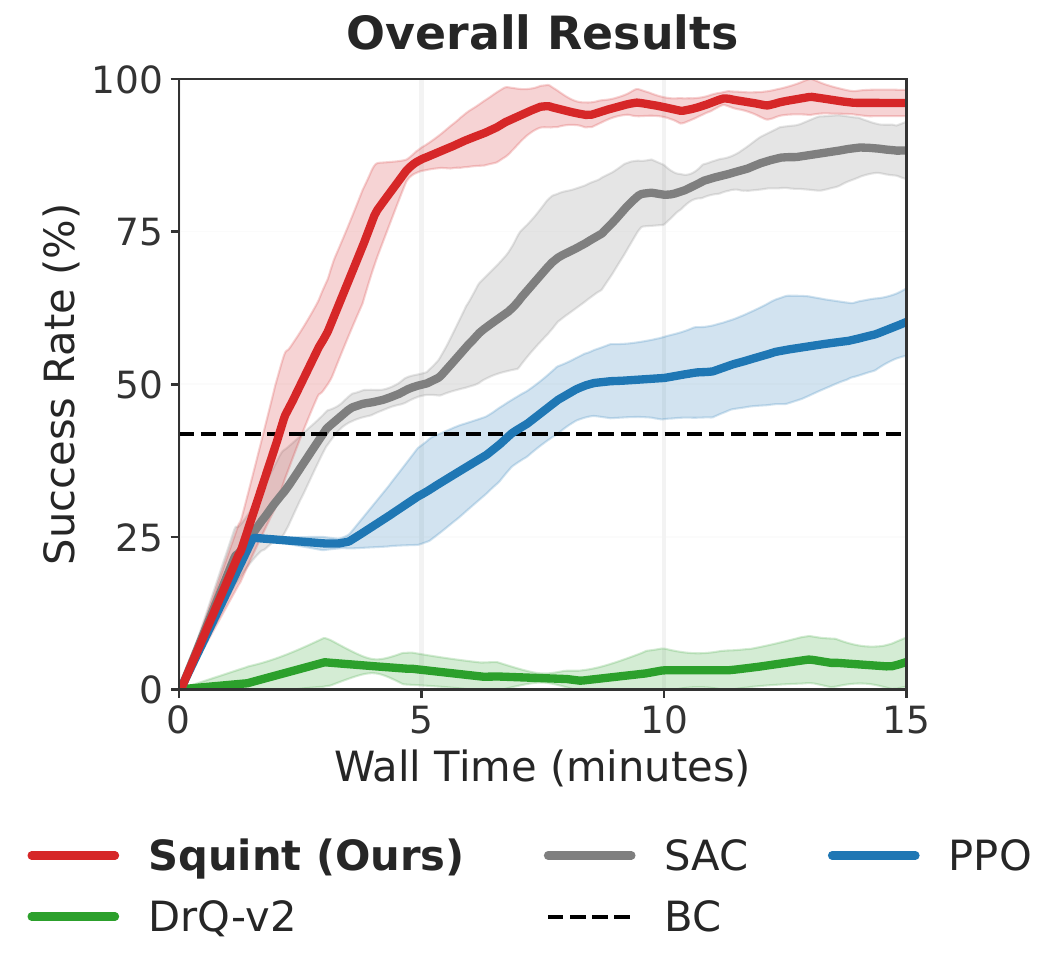}
      \end{minipage}
      \hfill
      \begin{minipage}{0.42\linewidth}
        \centering
        \includegraphics[width=\linewidth,trim={0cm 0.2cm 0cm 0.1cm},clip]{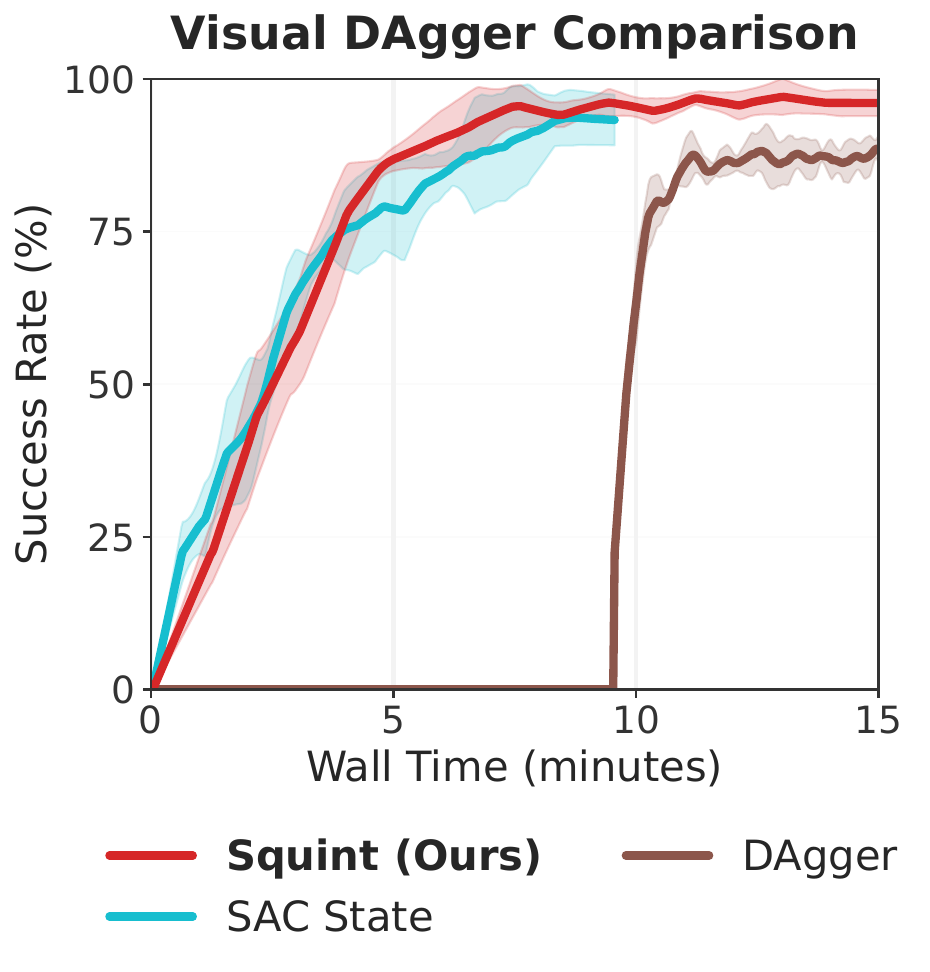}
      \end{minipage}
      \hspace{0.04\linewidth}
      }
    \refstepcounter{figure}
    \begin{flushleft}
    {\small Fig.~\thefigure. \textbf{Overall Results.} \textit{(Left)} Comparison with visual RL baselines. \textit{(Right)} Comparison with State-to-Visual DAgger, where we include the time it takes to train an SAC state expert into the comparison. Mean Success Rate and 95\% CI over 5 seeds on all tasks.}
    \end{flushleft}

    \vspace{-0.5cm}

  \label{fig:combined_results}
\end{figure*}

\textbf{Replay Buffer Size.} Due to our use of squinting and low image sizes of 16 x 16, we are able to scale replay buffer size to easily fit on the GPU for faster training speed. We compare using 100K vs 1M replay buffer storage size, and find that the 1M improves asymptotic method success rate on our task set by 7\%. While we use a simple FIFO buffer, we leave it to future work to use more advanced buffer sampling techniques \cite{schaul2015prioritized, hou2017novel} that could further improve wall time.

\textbf{Simulation Control Frequency.} To facilitate faster training times, we train with 10Hz control frequency using a PD Joint Position Target Delta controller. We use large position offset limits, to allow the agent to move faster in simulation.

\textbf{Real-World Robot Transfer.} To transfer trained agents to the real robot, without compromising safety, we scale all actions down by 0.15, and triple the control frequency to 30Hz. While this constitutes  a mismatch between simulation and real world control frequency, we find that a higher control frequency with a smaller action magnitude in the real world provides smoother trajectories.

\begin{figure*}[t]
    \centering
    \begin{minipage}[t]{0.48\textwidth}
        \vspace{0pt}
        \centering
        \label{tab:sim_results}
        \captionof{table}{Simulation Success Rates (\%)}
        \resizebox{\textwidth}{!}{\centering
\begin{tabular}{lccccc}
\toprule
\textbf{Task} & \textbf{Squint (Ours)} & \textbf{SAC} & \textbf{PPO} & \textbf{DrQ-v2} & \textbf{BC} \\
\midrule
Reach Cube & 100.0\tiny{$\pm$0.0} & 99.4\tiny{$\pm$1.1} & 88.0\tiny{$\pm$10.3} & 12.3\tiny{$\pm$8.3} & 61.2\tiny{$\pm$9.2} \\
Reach Can & 100.0\tiny{$\pm$0.0} & 99.1\tiny{$\pm$1.8} & 96.6\tiny{$\pm$2.4} & 23.6\tiny{$\pm$17.6} & 56.2\tiny{$\pm$12.5} \\
Lift Cube & 99.8\tiny{$\pm$0.4} & 98.1\tiny{$\pm$2.4} & 92.8\tiny{$\pm$4.5} & 0.0\tiny{$\pm$0.0} & 46.2\tiny{$\pm$3.1} \\
Lift Can & 98.8\tiny{$\pm$2.5} & 95.4\tiny{$\pm$3.5} & 96.2\tiny{$\pm$4.7} & 0.0\tiny{$\pm$0.0} & 53.8\tiny{$\pm$12.9} \\
Place Cube & 97.5\tiny{$\pm$3.1} & 99.1\tiny{$\pm$1.7} & 29.0\tiny{$\pm$33.8} & 0.0\tiny{$\pm$0.0} & 38.8\tiny{$\pm$8.3} \\
Place Can & 96.3\tiny{$\pm$3.0} & 98.6\tiny{$\pm$1.8} & 71.1\tiny{$\pm$18.3} & 0.0\tiny{$\pm$0.0} & 28.8\tiny{$\pm$8.5} \\
Stack Cube & 95.0\tiny{$\pm$6.1} & 97.7\tiny{$\pm$1.6} & 4.9\tiny{$\pm$9.7} & 0.0\tiny{$\pm$0.0} & 26.2\tiny{$\pm$9.2} \\
Stack Can & 81.2\tiny{$\pm$8.8} & 18.7\tiny{$\pm$23.2} & 3.0\tiny{$\pm$6.1} & 0.0\tiny{$\pm$0.0} & 23.8\tiny{$\pm$13.3} \\
\midrule
\textbf{All Tasks} & \textbf{96.1}\tiny{$\pm$1.6} & 88.3\tiny{$\pm$3.4} & 60.2\tiny{$\pm$4.0} & 4.5\tiny{$\pm$2.9} & 41.9\tiny{$\pm$4.2} \\
\bottomrule
\end{tabular}}
    \end{minipage}
    \hfill
    \begin{minipage}[t]{0.45\textwidth}
        \vspace{0pt}
        \centering
        \label{tab:real_results}
        \captionof{table}{Real-World Success Rates}
        \resizebox{\textwidth}{!}{\centering
\begin{tabular}{lccccc}
\toprule
\textbf{Task} & \textbf{Squint (Ours)} & \textbf{SAC} & \textbf{PPO} & \textbf{DrQ-v2} & \textbf{BC} \\
\midrule
Reach Cube & 10/10 & 10/10 & 7/10 & 4/10 & 8/10 \\
Reach Can & 10/10 & 10/10 & 10/10 & 4/10 & 7/10 \\
Lift Cube & 10/10 & 10/10 & 6/10 & 0/10 & 6/10 \\
Lift Can & 9/10 & 9/10 & 7/10 & 0/10 & 6/10 \\
Place Cube & 10/10 & 8/10 & 9/10 & 0/10 & 7/10 \\
Place Can & 10/10 & 6/10 & 4/10 & 0/10 & 0/10 \\
Stack Cube & 8/10 & 6/10 & 3/10 & 0/10 & 2/10 \\
Stack Can & 6/10 & 6/10 & 4/10 & 0/10 & 2/10 \\
\midrule
Total & 73/80 & 65/80 & 50/80 & 8/80 & 38/80 \\
\textbf{All Tasks} & \textbf{91.3\%} & 81.3\% & 62.5\% & 10.0\% & 47.5\% \\
\bottomrule
\end{tabular}}
    \end{minipage}
    \\[0.5cm]
    \includegraphics[width=1.0\linewidth,trim={0.cm 0.2cm 0.0cm 0.1cm},clip]{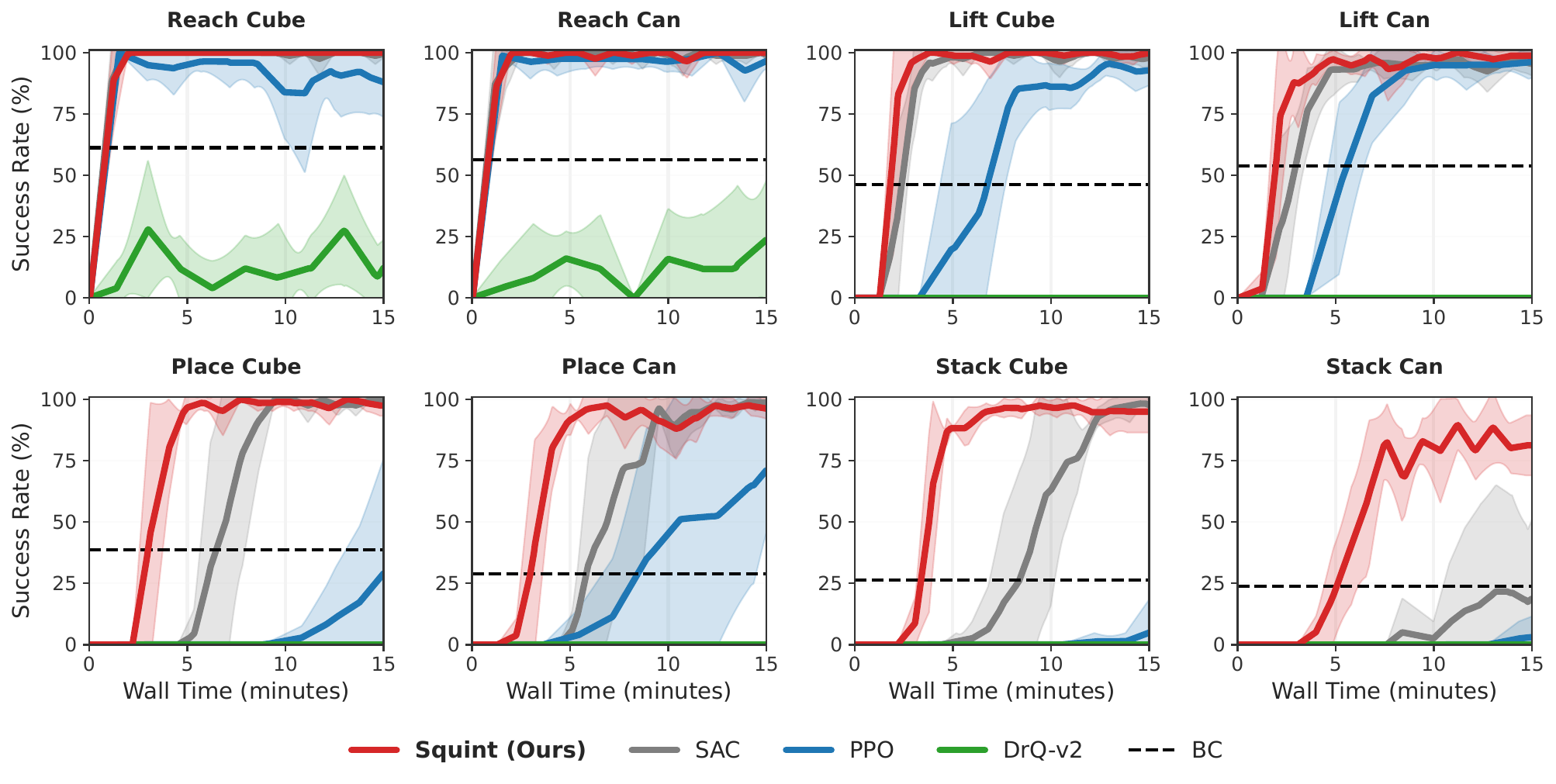}
    \vspace{-0.1cm}

    \refstepcounter{figure}
    \begin{flushleft}
    {\small Fig.~\thefigure. \textbf{Detailed Results.} We train all methods for 15 minutes on all 8 tasks, and plot the results above. Mean Success Rates and 95\% CI over 5 seeds. \textit{(Top Left)} Simulation success rate table of all methods after 15 minutes. \textit{(Top Right)} Real world success rate table over 10 trials for each task. \textit{(Bottom)} Per-task success rate plots in simulation during training.}
    \end{flushleft}

    \vspace{-0.1cm}

    \label{fig:detailed_results}
\end{figure*}

\section{Experiments}
\label{sec:exp}

We setup 8 distinct tasks in ManiSkill3 \cite{tao2024maniskill3} simulator with the 5-DoF SO-101 robotic arm \cite{cadene2024lerobot}, and train our method and baselines on each of these tasks for 15 minutes over 5 seeds. We only use the wrist camera image and proprioceptive state as input. We then deploy the best seed from each method on each task, and evaluate it zero-shot on our real world robot setup. Our real robot setup is shown in Figure \ref{fig:setup}.

\textbf{Task Design.} We design the tasks to be digital twins of the real world setup, thereby reducing the visual gap. We use the wrist camera for learning, as it has been shown to provide better representations than third person cameras in robot manipulation \cite{hsu2022vision, almuzairee2025merging}. For each task in our task set, the objects are randomly placed on the table in a 0.2m x 0.2m xy box around the gripper's starting xy position. The tasks are: \textit{(1) Reach Cube}: robot reaches 2 cm above the cube. \textit{(2) Reach Can}: robot reaches 2 cm above the can. \textit{(3) Lift Cube}: robot lifts the cube and moves to a predefined rest position. \textit{(4) Lift Can}: robot lifts the can and moves to a predefined rest position. \textit{(5) Place Cube}: robot lifts the cube and places it into the container. \textit{(6) Place Can}: robot lifts the can and places it into the container. \textit{(7) Stack Cube}: robot lifts the red cube and stacks it on top of the blue cube. \textit{(8) Stack Can}: robot lifts the cube and stacks it on top of the can. We show the start and success position of each task in Figure \ref{fig:header}.

\textbf{Domain Randomization Setup.} Even though we design simulation tasks to be similar to the real world setup, domain randomization is necessary to be able to bridge the gap between the simulation and the real world. We apply visual domain randomization consisting of small random position, rotation, and FOV perturbations to the wrist camera, as well as lighting and color jitter alterations. We apply physical domain randomization where we randomize object sizes, object frictions and gripper closing speed. Furthermore, we apply additive isotropic Gaussian noise $(\sigma = 5°)$ to joint positions in the proprioceptive state to ensure robustness to any joint position mismatches between simulation and real world setups. We \emph{emphasize} that Squint is able to achieve fast training speed despite these aggressive randomizations.  

\textbf{Baselines.} We compare our method, Squint, against the following competitive baselines: (1) \textbf{SAC}: an optimized implementation of Soft Actor Critic \cite{haarnoja2018soft} for parallel simulators, with Squint's tuned hyper parameters, but without Squint's architectural differences. (2) \textbf{PPO}: an optimized implementation of Proximal Policy Optimization \cite{schulman2017proximal}, the de facto standard for visual sim-to-real robotics due to its fast wall time training speed with parallel simulators. (3) \textbf{DrQ-v2}: Data Regularized Q-Learning, the sample efficiency standard for off-policy visual learning \cite{yarats2021mastering}. We note that we kept DrQ-v2 sequential with one environment, rather than parallelized, to stay true to its original implementation, and compare our method with the standard for off-policy visual learning. (4) \textbf{BC}: Behaviour Cloning \cite{pomerleau1988alvinn, atkeson1997robot} from expert demonstrations, commonly used for distilling visual agents. (5) \textbf{DAgger}: Dataset Aggregation \cite{ross2011reduction}, a stronger imitation learning baseline than Behaviour Cloning, commonly used to train visual agents when there is access to the environment and the expert agent. We specifically focus on State-to-Visual DAgger \cite{ross2011reduction, mu2025should}, where one trains an expert SAC state based agent and distills it into a visual actor for deployment.

\section{Results}

\textbf{Overall Results.} We display the \emph{overall average} plot in Figure \ref{fig:combined_results} and all \emph{detailed results} in Figure \ref{fig:detailed_results}, showcasing per-task breakdowns and the tables for both simulation and real world success rates. Our method, Squint, surpasses all baselines in terms of training wall-time speed, and achieves an average success rate of 96.1\% over all tasks at the end of 15 minutes of training. On the real world results, our method achieves 91.3\% success rate averaged over 80 trials for 8 tasks, surpassing baselines on the real robot setup. 

\begin{table}[t]
\centering
\caption{Real World Success - Compared with Visual DAgger.}
\label{tab:dagger_results}
\begin{tabular}{lccccc}
\toprule
 & \textbf{Squint (Ours)} & \textbf{Visual DAgger} \\
\midrule
Total & 73/80 & 53/80 \\
\textbf{Average} & \textbf{91.3\%} & 66.3\% \\
\bottomrule
\end{tabular}
\end{table}

\textbf{Comparison with Visual DAgger.} We further compare our method with Visual DAgger agents that are distilled from SAC state experts and display the results in Figure \ref{fig:combined_results}. We include the time it takes to train SAC state experts as done in \cite{mu2025should}.  We find that Squint achieves similar training speed as the SAC state agent, and achieves higher success rate at the end of training than the State-to-Visual DAgger agent on our task set. Furthermore, we deploy DAgger onto our real robot setup, showcasing the results in Table \ref{tab:dagger_results}, and we find that Squint outperforms DAgger's success rate by 25\% in the real world.

\section{Discussion}
\label{sec:discussion}

\textbf{Quantitative Analysis.} Based on simulation and real world results in Figure \ref{fig:detailed_results}, we find that the real world results closely align with the simulation results. This indicates that the simulated environments provide a good test bed for measuring success in the real world. Our SAC baseline outperforms the other baselines, but falls short of Squint, due to its architectural limitations. Our PPO baseline doesn't learn as fast as off-policy methods on the harder tasks of our task set, mainly due to the reuse of data that off-policy algorithms can leverage. Our DrQ-v2 baseline fails to learn, primarily due to it learning from one environment only with heavy randomization, and doesn't benefit from the speed of parallelization. Our BC baseline learns, but is limited in performance due to the state distribution mismatch that DAgger can alleviate. While DAgger learns better, we find that it still falls short of Squint. Both BC and DAgger share a key limitation in our setting, which is learning from a state-based agent. Our task setup uses wrist cameras, which require active vision, and a different exploration movement than an all-seeing state agent would take. Overall, Squint shows promise as it accelerates wall-time speed of training deployable visual agents to the real world.

\begin{figure*}[t]
    \centering
    \includegraphics[width=\linewidth]{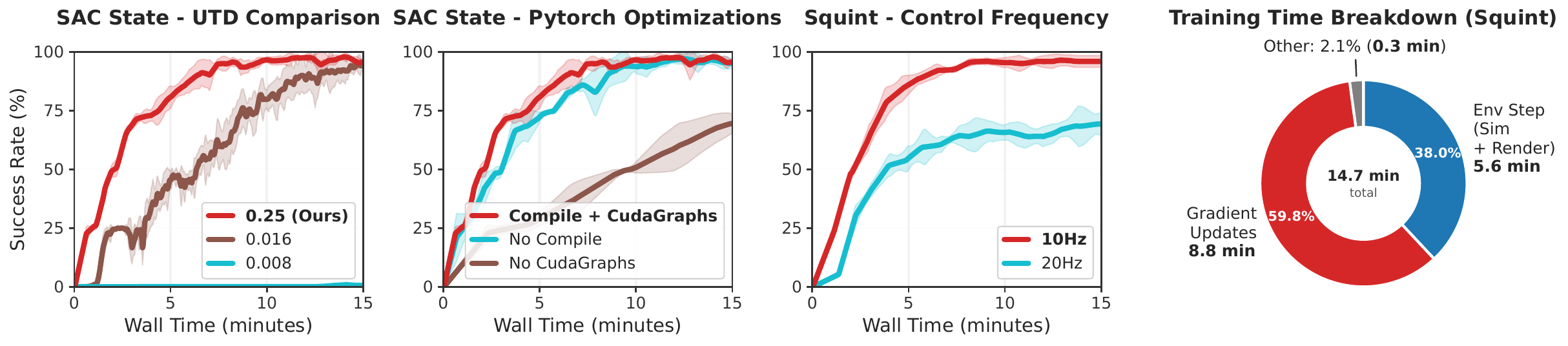}
    \refstepcounter{figure}
    \begin{flushleft}
    {\small Fig.~\thefigure. \textbf{Generality and Timing Analysis.} \textit{(Left)} We show that our UTD and Pytorch optimization choices are effective even for state-based RL. \textit{(Right)} We analyze simulation timing, where we alter Control Frequency in simulation to show its impact, and display a detailed Training Time Breakdown of Squint. Mean Success Rate and 95\% CI over 8 tasks and 5 seeds for all line plots.}
    \end{flushleft}
    \vspace{-0.5cm}
    \label{fig:extend}
\end{figure*}

\textbf{Generality and Timing Analysis.} We further analyze whether our design choices generalize to state-based RL, and the training simulation timing impact and breakdown in Figure \ref{fig:extend}. To deduce whether the high UTD benefits are mainly due to the visual RL pipeline, we train state based SAC agents with different UTD values, showing that a higher UTD is highly beneficial in manipulation domains. We also show that the Pytorch Optimizations are crucial for state based SAC, indicating that it is beneficial for both state and visual RL. We then show the impact of using a low control frequency in simulation of 10Hz, which allows faster training wall-clock time, as opposed to higher control frequencies. Finally, we display a detailed breakdown of Squint's training time, where 60\% of training time is dominated by the gradient updates, while 38\% of training time is spent on environment simulation and rendering.

\begin{table}[t]
\centering
\caption{Real World Success - Visual Robustness.}
\label{tab:jitter_results}
\begin{tabular}{lccccc}
\toprule
 & \textbf{Squint (Ours)} & \textbf{Squint - No Color Jitter} \\
\midrule
Total & 73/80 & 58/80 \\
\textbf{Average} & \textbf{91.3\%} & 72.5\% \\
\bottomrule
\end{tabular}
\end{table}

\textbf{Visual Robustness Analysis.} Despite building a digital twin of the environment in simulation, visual reinforcement learning agents are brittle to changes in visuals that they haven't encountered before \cite{almuzairee2024recipe}. We found that even slight lighting changes in the real robot setup were degrading the performance of our deployed agent. We ablate removing color jitter from our environment pipeline and evaluate it on the real robot setup. We display our results in Table \ref{tab:jitter_results}, where we find that our agent's real world performance decreases by 18\% on all tasks without the application of color jitter. We leave it to future work to integrate more visual robustness into the training pipeline.

\textbf{Camera Resolution Analysis. } In Table \ref{tab:res_table}, we trained a higher camera resolution variant of squint (32x32) compared to our original formulation (16x16) and found that the sim-to-real transfer on the real robot is similar. The more we increase the resolution, the longer it takes to train in simulation, without an added benefit to the real world success rate. Using high resolution for visual RL has long been a non-negotiable. In this work, we prove quiet the opposite, that RL can learn even on the most blurriest and smallest photos, and that with squinting, researchers get a free lunch with a faster visual RL pipeline.  We also empirically found that one can reduce the resolution further and train successfully in simulation but the policy fails to transfer to the real robot, mainly due to overfitting. We leave mitigating this overfitting to future work.

\section{Conclusion} 
\label{sec:conclusion}

\textbf{Summary.} Fast GPU-based parallel simulators have dramatically lowered the entry barrier for robot learning. In this work, we integrate the 5-DoF SO-101 robotic arm into ManiSkill3, establishing 8 benchmark tasks with extensive domain randomization for sim-to-real transfer. We introduce Squint, an off-policy visual reinforcement learning algorithm that, through careful architectural design, hyperparameter tuning and image preprocessing, achieves superior wall-time efficiency and success rates compared to existing baselines on our proposed SO-101 Task Set. Squint trains on a single RTX 3090 GPU in under 15 minutes, producing policies ready for real-world deployment. We hope this work accelerates iteration cycles in robot learning, makes the field more accessible to researchers, and provides a strong foundation for future advances in visual reinforcement learning.

\begin{table}[t]
\centering
\caption{Real World Success - Camera Resolutions.}
\label{tab:res_table}
\begin{tabular}{lccccc}
\toprule
 & \textbf{Squint 16x16 (Ours)} & \textbf{Squint 32x32} \\
\midrule
Total & 73/80 & 73/80 \\
\textbf{Average} & \textbf{91.3\%} & \textbf{91.3}\% \\
\bottomrule
\end{tabular}
\end{table}







\clearpage
\bibliographystyle{IEEEtran}
\bibliography{references}

\onecolumn
\section*{APPENDIX}
\vspace{1cm}

\textbf{Limitations and Opportunities.} Building on this work, there are several limitations and promising directions that warrant further investigation:
\vspace{0.2cm}
\begin{itemize}
    \item \textbf{Visual Robustness.} Visual reinforcement learning agents remains brittle to visual changes. While most prior works in off-policy learning focus on improving visual robustness under the sample efficiency axis, optimizing visual robustness under the wall-time axis presents an equally compelling objective—whether through visual augmentations \cite{hansen2021stabilizing, almuzairee2024recipe, teoh2024green, ma2025comprehensive}, pretrained encoders \cite{yuan2022pre, wang2023generalizable, brown2026segdac}, or auxiliary representation learning objectives \cite{batra2024zero, echchahed2025survey}.
    \item \textbf{Sample Efficiency.} Numerous advances in off-policy learning have yielded significant gains in sample efficiency \cite{hansen2024tdmpc2, lee2024simba, lee2025hyperspherical, fujimoto2025mrq, castanyer2025stable, obando2025simplicial, chang2026surprising}. Incorporating these methods into Squint while maximizing wall-time efficiency is a promising direction.
    \item \textbf{Gripper Design.} Despite the 5 DoF robot's remarkable capability for its price point, sim-to-real transfer for can grasping was challenging due to tipping and insufficient gripper friction. More adhesive \cite{glick2018soft} or alternative grip designs could mitigate this issue.
    \item \textbf{Generalization.} While this work addresses single-task visual RL, natural extensions include multi-task \cite{Espeholt2018IMPALASD, xu2024rldg, nauman2025brc}, multi-view \cite{yuan2024maniwhere, almuzairee2025merging, li2025manivid, kim2025seeing}, or multi-agent \cite{yu2022marlppo} settings as a step towards more generalist agents.
    \item \textbf{Privileged Training.} We formulated Squint as a symmetric visual based reinforcement learning agent for simplicity. Extending Squint to an asymmetric pipeline, where the critic could leverage privileged information \cite{pinto2017asymmetric, hu2024privileged}, could further speed up the learning.
    \item \textbf{Dataset Diversity.} Our work extensively evaluated the SO-101 task set in simulation and the real world. Evaluating and improving Squint on larger and diverse visual benchmarks is another promising direction \cite{tassa2018deepmind, yu2020meta, hansen2025newt}.
    \item \textbf{Imitation Learning Limitations.} Imitation learning agents achieve suboptimal policies when given an active vision observation space and asked to imitate a state based agent. Future work could introduce imitation learning agents that address this limitation \cite{hu2024privileged, song2025distill}.
    \item \textbf{Sim and Real Co-training.} While we train Squint policies purely from simulation, co-training with both parallel simulation and real world demonstrations could prove a powerful way to bootstrap the learning \cite{maddukuri2025sim}.  
\end{itemize}

\clearpage

\begin{figure*}[t]
    \centering
    
    \includegraphics[width=\linewidth]{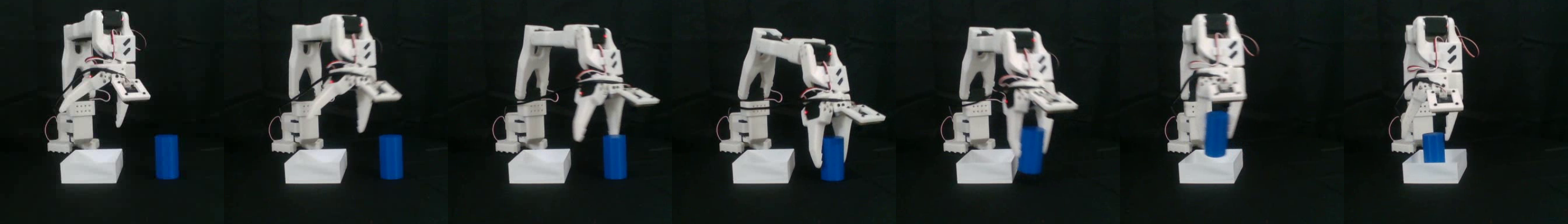}
    \vspace{-0.2cm}
    {\small (a) Place Can task.}
    \vspace{0.5cm}
    
    \includegraphics[width=\linewidth]{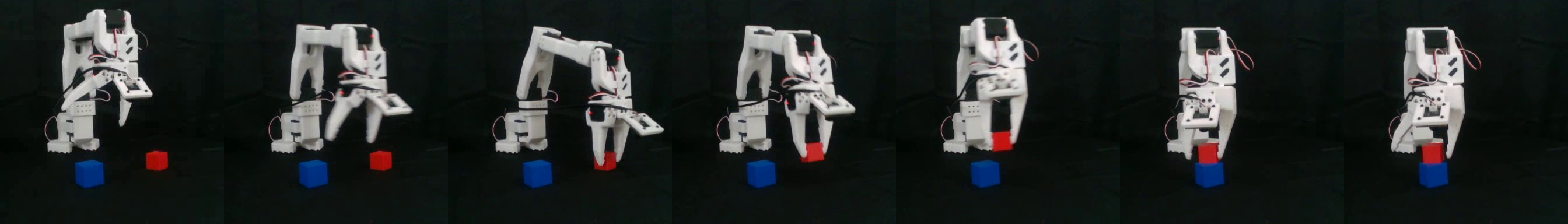}
    \vspace{-0.2cm}
    {\small (b) Stack Cube task.}
    \vspace{0.5cm}
    
    \includegraphics[width=\linewidth]{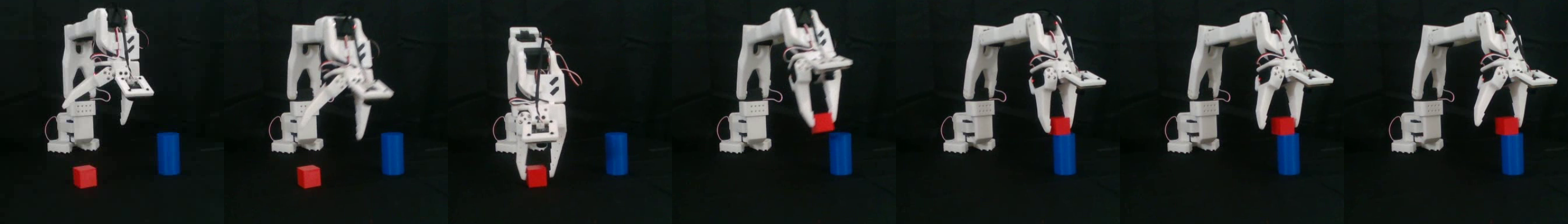}
    \vspace{-0.2cm}
    {\small (c) Stack Can task.}
    \vspace{0.2cm}

    \refstepcounter{figure}
    \begin{flushleft}
    {\small Fig.~\thefigure. \textbf{Qualitative Examples.} We show timesteps from left to right, and demonstrate successful deployment on our real world robot.}
    \end{flushleft}
    
    \label{fig:qual}
\end{figure*}


\textbf{Qualitative Analysis.} We display sample trajectories of Squint on the real SO-101 robot in Figure \ref{fig:qual}. The three trajectories consist of the three hardest tasks: PlaceCan, StackCube and StackCan.

\clearpage
\textbf{Algorithm Overview.} Algorithm \ref{alg:training} outlines the full training loop of Squint. The critic and encoder are updated by \texttt{num\_updates} gradient updates while the actor is updated less frequently by \texttt{(num\_updates / policy\_freq)} gradient updates. For full details, please refer to the open source code.

\renewcommand{\algorithmiccomment}[1]{\textsc{#1}}
\begin{algorithm*}[h]
\caption{{Squint: Pseudocode (distributional critic is omitted for simplicity)}}\label{alg:training}
\begin{algorithmic}[1]
\State Initialize encoder $f_\psi$, actor $\pi_{\phi}$, two critics $Q_{\theta_{1}}, Q_{\theta_{2}}$, entropy temperature $\alpha$, replay buffer $\mathcal{B}$
\State Initialize target critics $Q_{\bar{\theta}_{1}}, Q_{\bar{\theta}_{2}}$ with $\bar{\theta}_{1} \leftarrow \theta_{1}$ and $\bar{\theta}_{2} \leftarrow \theta_{2}$
\For{each environment step $t$} \vspace{0.01in}
\State Observe $(o_t^{\text{rgb}}, s_t^{\text{proprio}})$ and downsample image $\tilde{o}_t = \textsc{Downsample}(o_t^{\text{rgb}})$
\State Sample $a_t \sim \pi_{\phi}(f_\psi(\tilde{o}_t), s_t^{\text{proprio}})$ and take action $a_t$
\State Observe next $(o_{t+1}^{\text{rgb}}, s_{t+1}^{\text{proprio}})$ and reward $r_t$
\State Store transition $\tau = (\tilde{o}_t, s_t^{\text{proprio}}, a_t, r_t, \textsc{Downsample}(o_{t+1}^{\text{rgb}}), s_{t+1}^{\text{proprio}})$ in $\mathcal{B} \leftarrow \mathcal{B} \cup \{\tau\}$
\For{$j=1$ {\bfseries to} $\texttt{num\_updates}$}
\State Sample mini-batch $B =\{\tau_{k}\}_{k=1}^{|B|}$ from $\mathcal{B}$
\State Encode observations: $z_t \gets f_\psi(\tilde{o}_t)$, \quad $z_{t+1} \gets f_\psi(\tilde{o}_{t+1})$
\State \textbf{Compute target Q-value} via average:
\State \;\; $\tilde{a}_{t+1} \sim \pi_{\phi}(\cdot|z_{t+1}, s_{t+1}^{\text{proprio}})$
\State \;\; $y = r_t + \dfrac{\gamma}{2}\displaystyle\sum_{i=1}^{2}\left(Q_{\bar{\theta}_{i}}(z_{t+1}, s_{t+1}^{\text{proprio}}, \tilde{a}_{t+1}) - \alpha \log \pi_{\phi}(\tilde{a}_{t+1}|z_{t+1}, s_{t+1}^{\text{proprio}}) \right)$
\State \textbf{Update critic and encoder}:
\State \;\; $(\psi, \theta_{i}) \leftarrow (\psi, \theta_{i}) - \nabla_{(\psi, \theta_{i})}\dfrac{1}{|B|}\displaystyle\sum_{\tau_{k} \in B}\Big(Q_{\theta_{i}}(z_t, s_t^{\text{proprio}}, a_t) - y\Big)^{2}$ for $i \in \{1, 2\}$
\State \textbf{Update entropy temperature:}
\State \;\; $\alpha \leftarrow \alpha - \nabla_{\alpha} \dfrac{1}{|B|}\displaystyle\sum_{\tau_{k} \in B} (\mathcal{H}^{\texttt{target}}  - \mathcal{H}(z_t, s_t^{\text{proprio}})) \cdot \alpha$
\If{$j \;\%\; \texttt{policy\_freq} == 0$}
\State \textbf{Update actor}:
\State \;\; $z_t^{\text{sg}} \gets \textsc{stopgrad}(z_t)$
\State \;\; $\tilde{a}_t \sim \pi_{\phi}(\cdot |z_t^{\text{sg}}, s_t^{\text{proprio}})$
\State \;\; $\phi \leftarrow \phi + \nabla_{\phi}\dfrac{1}{2|B|}\displaystyle\sum_{\tau_{k} \in B}\sum_{i=1}^{2}\Big(Q_{\theta_{i}}(z_t^{\text{sg}}, s_t^{\text{proprio}}, \tilde{a}_t) - \alpha \log \pi_{\phi}(\tilde{a}_t|z_t^{\text{sg}}, s_t^{\text{proprio}})\Big)$
\EndIf
\State \textbf{Update target critic} $\bar{\theta}_{i} \leftarrow \rho \bar{\theta}_{i} + (1-\rho)\theta_{i}$ for $i \in \{1, 2\}$
\EndFor
\EndFor
\end{algorithmic}
\end{algorithm*}

\clearpage

\end{document}